%% file: main.tex
\documentclass[9pt, conference]{IEEEtran}
\IEEEoverridecommandlockouts
\usepackage{cite}
\usepackage{amsmath,amssymb,amsfonts}
\usepackage{graphicx}
\usepackage{textcomp}
\usepackage{xcolor}

\usepackage{nccmath}  
\usepackage{bbm}
\usepackage{mathtools}

\usepackage{amsfonts}
\usepackage{gensymb}

\usepackage{multirow}
\usepackage{enumitem}
\usepackage{xcolor}

\usepackage{booktabs}

\usepackage[titlenumbered,ruled]{algorithm2e}
\usepackage{algorithmicx}
\makeatletter
\newcommand{\removelatexerror}{\let\@latex@error\@gobble}
\makeatother

\usepackage{pifont}
\newcommand\mynuma[1]{\ifcase#1 \or \ding{172}\or \ding{173}\or
  \ding{174}\or \ding{175}\or \ding{176}\or \ding{177}%
  \or \ding{178}\or \ding{179}\or \ding{180}\or \ding{181}\else *\fi\relax}

\newcommand\mynumb[1]{\ifcase#1 \or \ding{182}\or \ding{183}\or
  \ding{184}\or \ding{185}\or \ding{186}\or \ding{187}%
  \or \ding{188}\or \ding{189}\or \ding{190}\or \ding{191}\else *\fi\relax}

\definecolor{Note_color}{rgb}{0.0, 0.0, 0.0}

\definecolor{Note_color2}{rgb}{1.0, 0.0, 0.0}

\definecolor{assign_color}{rgb}{1.0, 0.0, 0.0}

\definecolor{figure_color}{rgb}{0.0, 0.5, 0.0}

\def\BibTeX{{\rm B\kern-.05em{\sc i\kern-.025em b}\kern-.08em
    T\kern-.1667em\lower.7ex\hbox{E}\kern-.125emX}}
\begin{document}

\title{GPT4AIGChip: Towards Next-Generation AI Accelerator Design Automation via Large Language Models}

\author{
\IEEEauthorblockN{Yonggan Fu$^\ddag$, Yongan Zhang$^\ddag$, Zhongzhi Yu$^\ddag$, Sixu Li, Zhifan Ye, Chaojian Li, Cheng Wan, Yingyan (Celine) Lin}
\IEEEauthorblockA{\textit{School of Computer Science, Georgia Institute of Technology}}
\IEEEauthorblockA{\textit{\{yfu314, yzhang919, zyu401, sli941, zye327. cli851, cwan39, celine.lin\}@gatech.edu}}
\vspace{-1.5em}
}

\maketitle

\def\thefootnote{$\ddag$}\footnotetext{Equal contribution.}

\input{Sections/0-Abstract}

\input{Sections/1-Introduction}

\input{Sections/2-Examination}

\input{Sections/3-Framework}

\input{Sections/4-Experiment}

\input{Sections/5-Envision}

\input{Sections/6-Related-Work}

\input{Sections/7-Conclusion}

\bibliographystyle{IEEEtranS}
\bibliography{ref}

\end{document}

%% file: Sections/0-Abstract.tex
\begin{abstract}

The remarkable capabilities and intricate nature of Artificial Intelligence (AI) have dramatically escalated the imperative for specialized AI accelerators. Nonetheless, designing these accelerators for various AI workloads remains both labor- and time-intensive. While existing design exploration and automation tools can partially alleviate the need for extensive human involvement, they still demand substantial hardware expertise, posing a barrier to non-experts and stifling AI accelerator development. Motivated by the astonishing potential of large language models (LLMs) for generating high-quality content in response to human language instructions, we embark on this work to examine the possibility of harnessing LLMs to automate AI accelerator design. Through this endeavor, we develop GPT4AIGChip, a framework intended to democratize AI accelerator design by leveraging human natural languages instead of domain-specific languages. Specifically, we first perform an in-depth investigation into LLMs' limitations and capabilities for AI accelerator design, thus aiding our understanding of our current position and garnering insights into LLM-powered automated AI accelerator design. Furthermore, drawing inspiration from the above insights, we develop a framework called GPT4AIGChip, which features an automated demo-augmented prompt-generation pipeline utilizing in-context learning to guide LLMs towards creating high-quality AI accelerator design. To our knowledge, this work is the first to demonstrate an effective pipeline for LLM-powered automated AI accelerator generation. Accordingly, we anticipate that our insights and framework can serve as a catalyst for innovations in next-generation LLM-powered design automation tools.

\end{abstract}


%% file: Sections/1-Introduction.tex
\vspace{-0.2em}
\section{Introduction }
\vspace{-0.2em}

The landscape of artificial intelligence (AI), driven by deep neural networks (DNNs), has recently undergone transformational progress, leading to a pressing demand for specialized AI accelerators. The remarkable capabilities and the intricate nature of AI workloads have further amplified this demand.
However, the design of specialized accelerators catering to diverse AI tasks remains an arduous and time-consuming venture. 
Moreover, the level of hardware expertise required for using existing design exploration and automation tools \cite{Xu_2020, jia2021tensorlib, venkatesan2019magnet,zhang2018dnnbuilder,guan2017fp,garg2021taxonomy, liang2020deepburning} presents a formidable challenge for non-experts, stifling the innovative advancement in AI accelerators. This complex and technical domain is currently characterized by a steep learning curve, which limits access and restricts the expansion of AI accelerator design to general AI developers, creating an increasing gap between the pace of AI algorithm development and corresponding accelerators. 

In the face of these challenges, we find inspiration in the emerging capabilities of large language models (LLMs)~\cite{koubaa2023gpt,ouyang2022training,touvron2023llama,zhang2022opt}, with their amazing capacity to generate high-quality content based on human language instructions. These capabilities present a tantalizing prospect, inspiring the central question of this study: ``\textit{Can we harness the power of LLMs to automate the design of AI accelerators}?" 
Specifically, as shown in Fig.~\ref{fig:general_pipeline}, LLM-powered AI accelerator design automation aims to explore the accelerator design space with the assistance of LLMs, thus generating high-quality accelerator implementation that can satisfy user requirements while at the same time minimizing human involvement.
To answer the above question, we first conduct a comprehensive investigation into the limitations and capabilities of LLMs, in terms of generating AI accelerator designs. This is to understand the current landscape, while also exploring how we can better harness the power of LLMs to automate AI accelerator design. Informed by the insights drawn from this investigation, we develop a framework dubbed GPT4AIGChip, denoting ``GPT for \textbf{AI} \textbf{G}enerated \textbf{C}hip". GPT4AIGChip aims to democratize AI accelerator design and thus make AI accelerator design more accessible, particularly to those not well-versed in hardware, by leveraging human natural language as design instructions rather than relying on domain-specific languages. 
We summarize our contributions as follows:

\begin{figure}[t]
    \centering
    \includegraphics[width=0.95\linewidth]{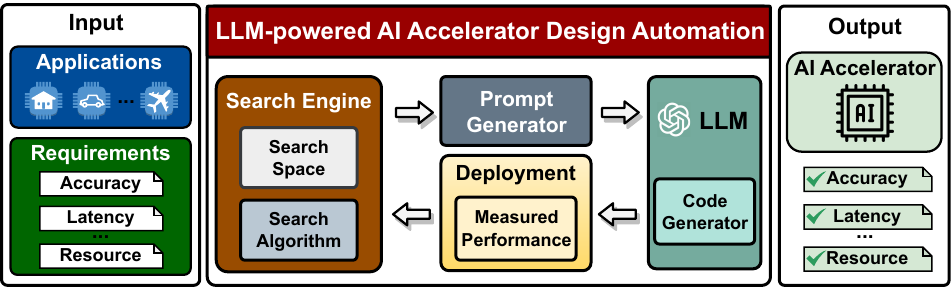}
    \vspace{-0.6em}
    \caption{A generic LLM-powered AI accelerator design automation pipeline.}
    \label{fig:general_pipeline}
    \vspace{-1.8em}
\end{figure}

\begin{figure*}[t]
    \centering
    \vspace{-1em}
    \includegraphics[width=\linewidth]{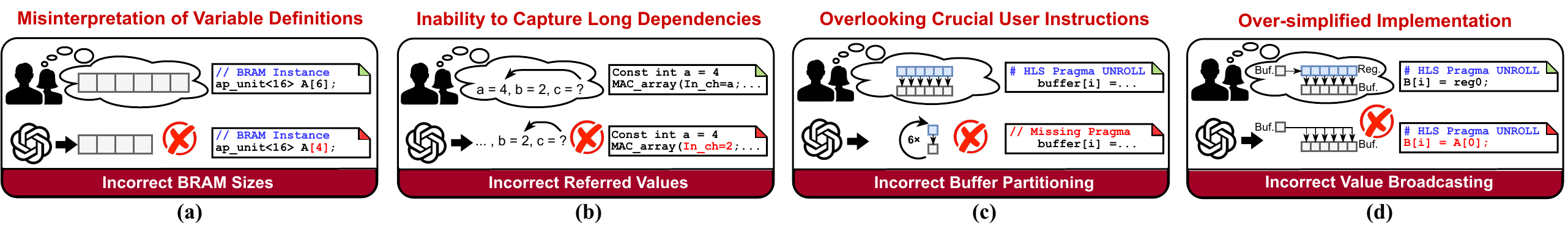}
    \vspace{-2em}
    \caption{Visualization of the identified common failures and limitations of existing LLMs for AI accelerator design automation.}
    \vspace{-1.3em}
    \label{fig:fails}
\end{figure*}

\vspace{-0.2em}
\begin{itemize}
    \item We thoroughly investigate the limitations and capabilities of leveraging existing LLMs to generate AI accelerator designs in order to understand our current position and draw useful insights on how to effectively leverage current LLMs in a design automation pipeline. As a tangible application of these insights, we've developed GPT4AIGChip, a framework that is the first to demonstrate LLM-powered AI accelerator design automation.

    \item Through the above comprehensive investigation, we have identified three crucial insights for leveraging the strengths of current LLMs: \underline{Insight-(1)} current LLMs struggle with understanding lengthy codes exhibiting long dependencies, particularly for their infrequently seen languages like high-level synthesis (HLS), thus necessitating the decoupling of different hardware functionalities in the design space; \underline{Insight-(2)} given the scarcity of annotated data for efficient finetuning of an open-sourced LLM, employing a mix of in-context learning and the logical reasoning prowess of typically closed-sourced yet powerful LLMs is a more effective choice; and \underline{Insight-(3)} it is critical to augment the prompts of LLMs with high-quality demonstrations, which are correlated to the context of the input design instructions.

    \item Our GPT4AIGChip instantiates the aforementioned Insight-(1) by constructing a decoupled accelerator design template written in HLS. In this way, it decouples different hardware modules and functionalities of an accelerator design, thus for the first time enabling LLM-powered AI accelerator design automation. 

    \item We instantiate the aforementioned Insight-(2)/-(3) by equipping GPT4AIGChip  with a demo-augmented prompt generator, which leverages both the in-context learning and the logical reasoning capabilities of powerful GPT models, enabling automated exploration of the accelerator search space powered by LLMs.

    \item Extensive experiments validate and demonstrate the effectiveness of our GPT4AIGChip framework in generating high-quality AI accelerator design, while significantly reducing the required human efforts and expertise in the design process.
    
\end{itemize}

The insights and framework derived from this work can spark further innovation in next-generation LLM-powered design automation tools and illuminate the promising field of AI-generated accelerators.

%% file: Sections/2-Examination.tex
\section{\textbf{Where We Are}: The Limitations and Capabilities of Current LLMs for AI Accelerator Design}
\label{sec:examination}

\subsection{An Overview of the Assessment}
\label{sec:motivation_for_assessment}
\vspace{-0.3em}

While LLMs excel in various generation tasks (e.g., question answering, language translation, and chatbot dialogues), these tasks mainly involve natural languages, which LLMs are extensively trained on. However, their ability to handle languages and tasks that they encounter less frequently during pretraining, e.g., generating AI accelerator designs using HLS languages, remains an open question. Therefore, to effectively utilize LLMs in automating AI accelerator design, it is crucial to have a comprehensive understanding of the capabilities and limitations of state-of-the-art (SOTA) LLMs. This could help avoid unwarranted optimism or pessimism. Our assessment aims to provide this understanding, serving as a foundation for future innovations in LLM-powered automated AI accelerator design.

To attain this goal, we first identify the common limitations of existing LLMs for AI accelerator design and then validate whether finetuning open-sourced LLMs on annotated hardware codes (e.g., HLS) could enhance their understanding of hardware codes and design. Ultimately, considering the identified shortcomings in these two steps, we reconsider the capabilities of LLMs that could be effectively utilized for practical AI accelerator design automation.

\vspace{-1em}
\subsection{Failures and Limitations of Existing LLMs}
\label{sec:failures}
\vspace{-0.3em}

To leverage LLMs in generating HLS implementations of AI accelerators based on user instructions, one intuitive approach is to pair user instructions with commonly adopted HLS templates to serve as LLM prompts. In this subsection, we adapt an HLS implementation from~\cite{zhang2020skynet} as our template, which utilizes a widely accepted for-loop-based design approach. We assess the capabilities of the SOTA LLM, GPT-4~\cite{gpt4}, in generating hardware implementations according to our specified instructions. However, we observe that LLMs frequently generate non-synthesizable, functionally incorrect code. The common failures are outlined below.

\textbf{Misinterpretation of variable definitions.}
Implementing hardware accelerator code requires precise definitions of variables or functions to accurately instantiate hardware modules. However, LLMs often struggle with this task. For instance, accurately creating an array variable to match the size of an on-chip Block RAM (BRAM) is crucial for computation tile sizes and data reuse levels. Unfortunately, as illustrated in Fig.~\ref{fig:fails} (a), LLMs may instantiate modules with incorrect sizes, even when following given instructions. This issue arises from LLMs' difficulty in understanding the connection between identifiers in the instruction and the variables defined in the code, leading to misinterpretation of the provided quantities in the instruction.

\textbf{Inability to capture long dependencies.} AI accelerator parameters are often intertwined, where a set of parameters for one module can influence and be influenced by other modules, causing long dependencies. Current LLMs frequently falter when handling these long dependencies, neglecting earlier design configurations when generating later modules. For example, as mentioned earlier, BRAM sizes can affect tiled computation sizes, which may in turn influence the number of processing elements (PEs). However, as Fig.~\ref{fig:fails} (b) shows, when we specify the parallelism factor along the input channel dimension of weights to be $4$, LLMs disregard this when generating MAC arrays and consider only $2$ parallel input channels.

\textbf{Overlooking crucial user instructions.} LLMs may struggle with reasoning about the relationships (1) among different hardware design concepts and (2) between natural language instructions and the code to be generated, thus overlooking crucial user instructions. Unlike the above long-dependency issue, this limitation involves relationships among multiple design concepts, which are not necessarily separated by long code blocks. For example, \textit{unrolling (parallelizing)} computation along certain data dimensions necessitates \textit{partitioning} the BRAM instance of the dependent data along the same dimensions. However, as Fig.~\ref{fig:fails} (c) shows, the code generated by existing LLMs often treats the design concepts in isolation. Additionally, LLMs may struggle to link natural language instructions with appropriate code generation, e.g., when instructed to modularize certain sub-modules, they may fail to identify the relevant domain-specific pragmas.

\textbf{Over-simplified implementation.}
AI accelerator implementation requires certain coding styles (e.g., HLS) to accurately depict hardware behaviors. For instance, to broadcast a single element from one buffer to multiple locations in another (partitioned) buffer, direct value assignment between the two buffers should be avoided. Instead, we need to first assign the element to a register-like variable and then assign the data from this intermediate register to multiple locations in the other buffer, thus preventing access conflicts within the source buffer. However, as shown in Fig.~\ref{fig:fails} (d), existing LLMs often overlook such details in hardware design and generate impractical designs.

\begin{table}[t]
    \centering
    \vspace{-0em}
   \caption{Benchmark the designs generated by prompting (1) the closed-sourced GPT-4, (2) the open-sourced CodeGen~\cite{nijkamp2022codegen} \textit{w/o} finetuning, and (3) the finetuned CodeGen~\cite{nijkamp2022codegen}.}
    \vspace{-0.3em}
    \resizebox{\linewidth}{!}{
    \begin{tabular}{c|ccc}
        \toprule
       LLM & GPT-4 w/o finetune & CodeGen w/o finetune & CodeGen w/ finetune \\
       \midrule
       Pass@100 & 42\% & 0\% & 31\% \\
       \bottomrule
    \end{tabular}
    }
    \vspace{-1.5em}
    \label{tab:which_to_choose}
\end{table}

\vspace{-0.3em}
\subsection{Closed-sourced LLMs vs. Finetuned Open-sourced LLMs}
\label{sec:finetuning}
\vspace{-0.3em}

The identified limitations above indicate LLMs' deficiency in understanding hardware design codes like HLS. One potential solution could be to finetune open-sourced LLMs with annotated HLS codes. However, this approach faces challenges due to the lack of high-quality HLS codes annotated with corresponding design descriptions and closed-sourced advanced LLMs, preventing finetuning. These hurdles suggest adopting an open-sourced yet less powerful LLM. This subsection will investigate powerful closed-sourced LLMs versus finetuned open-sourced LLMs, aiming for an affordable solution without an overly large annotated dataset.

We benchmark the closed-sourced LLM, GPT-4~\cite{gpt4}, and the open-sourced LLM for code generation, CodeGen~\cite{nijkamp2022codegen}. Following the settings in~\cite{thakur2022benchmarking}, we target a fundamental AI accelerator design automation task: implementing the inner product of two vectors in HLS. We use the Pass@\textit{k} metric, representing the portion of successful compilations in \textit{k} attempts, as the performance metric. The prompt for both pipelines is: \textit{``Implement the inner product operator between two vectors in HLS. You are an expert in AI accelerator design with extensive HLS coding knowledge."}

\textbf{Finetuning strategy.} We adopt a two-stage finetuning process: First, we collect seven thousand open-sourced HLS code snippets from GitHub and finetune CodeGen~\cite{nijkamp2022codegen} using a masked prediction objective~\cite{devlin2018bert}. This enhances the LLM's hardware knowledge, particularly HLS implementations. Second, we create ten customized HLS templates with implementation instructions and finetune the LLM to generate the corresponding AI accelerator design. This equips the LLM with AI accelerator implementation expertise.

\textbf{Observations.} Tab.~\ref{tab:which_to_choose} indicates that the pretrained open-sourced CodeGen lacks proficiency in AI accelerator design without additional finetuning, as evident from its 0\% Pass@100 rate. Although our two-step finetuning process improves the model's capabilities, there remains a notable discrepancy (e.g., 11\% less Pass@100) compared to the closed-sourced GPT-4. These findings suggest that in data-limited situations, LLMs' initial competence in the target task outweighs their finetuning ability. Consequently, a powerful closed-sourced LLM becomes a more suitable choice.

 \vspace{-0.4em}
\subsection{Capabilities of LLMs}
\label{sec:mastered_abilities}
 \vspace{-0.3em}

Given the identified failures of LLMs, we reconsider their capabilities in practical LLM-powered AI accelerator design automation. Previous works~\cite{zheng2023codegeex,nijkamp2022codegen,koubaa2023gpt,chen2021evaluating,bubeck2023sparks} suggest that LLMs have learned rich representations that can be generalized to unseen coding problems through supervised finetuning or in-context learning~\cite{razeghi2022impact,olsson2022context,chan2022data}. To address the challenge of collecting high-quality instruction-code pairs required for finetuning, we explore the possibility of in-context learning by providing LLMs with HLS demonstrations composed of code pairs and corresponding design descriptions, in addition to user instructions. Inspired by the failure cases in Sec.~\ref{sec:failures}, we adopt decoupled hardware templates to avoid long dependencies, as detailed in Sec.~\ref{sec:llm_friendly_hw_template}. We summarize the identified capabilities of LLMs for successful AI accelerator design automation as follows:

\begin{figure}[t]
    \centering
    \vspace{-1.5em}
\includegraphics[width=\linewidth]{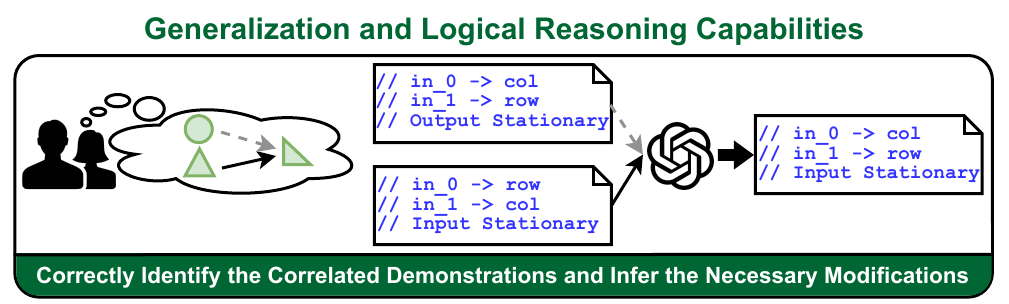}
    \vspace{-2em}
    \caption{Visualization of the identified generalization and logical reasoning capabilities of existing LLMs for AI accelerator design automation.}
    \label{fig:success}
    \vspace{-1.5em}
\end{figure}

\textbf{Generalization capability from in-context demonstrations.}
As shown in Fig.~\ref{fig:success}, given two snippets of HLS demonstrations, LLMs can generate a new hardware design following user instructions. They do so by maintaining the same coding style and modifying the design to meet new requirements, inferring the differences between the demonstrated design description and the user instruction. Our finding is that when high-quality demonstrations, which bear a certain degree of correlation to user instructions, are provided, the in-context learning capability of LLMs can be effectively activated.

\textbf{Logical reasoning capability under multiple demonstrations.}
Feeding more diverse demonstrations to LLMs could better harness their in-context learning capabilities. However, this necessitates logical reasoning skills to discern which demonstration should be referred to when implementing each item in the user instructions. As shown in Fig.~\ref{fig:success} (and in Sec.~\ref{sec:code_visual}), when given two different demonstrations, LLMs can provide a step-by-step analysis of the similarity between the functionalities required by the user instruction and those implemented in the demonstrations, subsequently selecting the appropriate one as a starting point. This reflects the logical reasoning capabilities of LLMs when provided with suitable demonstrations and also underscores the significance of high-quality demonstrations.

\section{
\textbf{What We Can Learn}: Insights from the Assessment}
\label{sec:insights}

Drawing upon our comprehension of the limitations and exploitable capabilities of LLMs in Sec.~\ref{sec:examination}, we distill insights and provide guidelines on how to efficiently harness LLMs for AI accelerator design automation. These insights establish the groundwork for our GPT4AIGChip framework in Sec.~\ref{sec:framework}. Moreover, they possess the potential to inspire future innovations in LLM-driven hardware design automation. In particular, the following insights are derived:

\textbf{Decoupled hardware template design.} Current LLMs struggle to generate effective AI accelerator designs when built upon commonly adopted HLS templates with long dependencies. This issue arises from the domain gap between the pretraining data and the infrequently encountered AI accelerator design tasks during pretraining, causing a lack of knowledge about both domain-specific languages and hardware functionality. Hence, it is crucial to \underline{(1)} properly design the hardware template to decouple different hardware functionalities in the design space, thus avoiding long dependencies, and \underline{(2)} decompose a complex task into simpler subtasks, i.e., different hardware functionalities should be generated independently and progressively, rather than generating lengthy codes all at once.

\textbf{Prioritize in-context learning given limited data.} Given the scarcity of annotated data, which is too limited to finetune an open-sourced but less powerful LLM in a supervised manner, leveraging the combination of in-context learning and logical reasoning capabilities of a closed-sourced but powerful LLM is a more effective choice.

\textbf{Proper prompt engineering.} The key to successful in-context learning for hardware design is augmenting prompts with high-quality demonstrations in the target domain. These demonstrations are anticipated to: \underline{(1)} exhibit a strong correlation with the input design instructions, and \underline{(2)} encompass a wide range of design parameters to convey ample domain knowledge, enhancing the capability of addressing various input instructions.

%% file: Sections/3-Framework.tex
\begin{figure}[t]
    \centering
    \vspace{-1em}
    \includegraphics[width=1.0\linewidth]{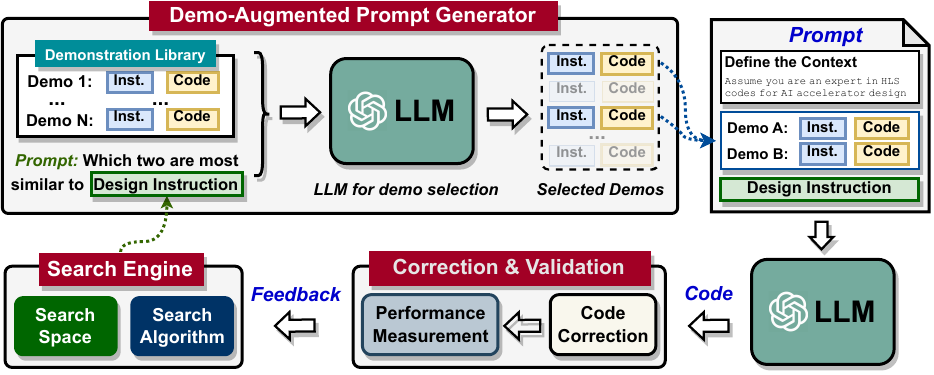}
    \vspace{-1.8em}
    \caption{Visualize the workflow of our proposed GPT4AIGChip framework.}
    \label{fig:framework}
    \vspace{-1.5em}
\end{figure}

\section{
\textbf{Instantiate the Drawn Insights}: The Proposed GPT4AIGChip Framework}

\label{sec:framework}

\vspace{-0.3em}
\subsection{The Overall Pipeline of GPT4AIGChip}
\label{sec:pipeline}
 \vspace{-0.3em}

Utilizing the insights from Sec.~\ref{sec:insights}, we develop the GPT4AIGChip framework, which is designed to empower non-expert users in leveraging LLMs for automated AI accelerator design. Notably, our GPT4AIGChip instantiates the three insights outlined in Sec.~\ref{sec:insights} by implementing in-context learning atop the closed-sourced but powerful GPT-4~\cite{gpt4} and incorporating two essential components: \underline{(1)} the LLM-friendly hardware template (see Sec.~\ref{sec:llm_friendly_hw_template}), which simplifies intricate AI accelerator codes into a modular structure, and \underline{(2)}
the demo-augmented prompt generator (see Sec.~\ref{sec:prompt_design}), which enhances LLMs' capacity to generate optimized AI accelerator designs by supplementing prompts with well-chosen demonstrations.

By integrating the LLM-friendly hardware template with the demo-augmented prompt generator, our GPT4AIGChip adopts an iterative approach to enhance the generated AI accelerator design, progressively approaching the optimal solution. Each iteration follows a four-stage workflow, as depicted in Fig.~\ref{fig:framework} and outlined below:

\begin{itemize}
    \item The search engine identifies the next design and its corresponding instruction for each module in the LLM-friendly hardware template, drawing on feedback from previously searched designs to guide the implementation and evaluation.
    \item The demo-augmented prompt generator creates a prompt for each module, combining relevant demonstrations (instruction-code pairs) to enhance the LLMs' in-context learning.
    \item The LLMs equipped with the above prompts sequentially generate the hardware design implementation.
    \item The design validation flow scrutinizes LLMs' generated codes, executing necessary modifications to assure deployability. 
\end{itemize}

\vspace{-0.3em}
\subsection{The LLM-friendly Hardware Template Design}
\label{sec:llm_friendly_hw_template}
\vspace{-0.3em}

As highlighted in Sec.~\ref{sec:insights}, providing the LLM with a hardware design template is vital to compensate for its limited AI accelerator design knowledge. Yet, as shown in Sec.~\ref{sec:failures}, existing HLS accelerator templates~\cite{qin2022enabling,wei2023hlsdataset, zhang2020skynet,fu2021auto} pose significant challenges for LLM-based AI accelerator generation due to their complex design parameter coupling and inter-dependency. To address this, we first establish design principles for an LLM-friendly accelerator micro-architecture and source code template. Guided by these principles, we propose a unique modular AI accelerator template, tailored to optimize LLM's capabilities in generating AI accelerator designs. We then discuss the implications and advantages of our template in enhancing LLM-assisted design generation across broader scenarios.

\textbf{The desired template design principles.}
To ensure effective LLM-assisted generation of accelerator designs, we identify three key principles for design templates: (1) high modularity, (2) decoupled module design, and (3) deep design hierarchies to facilitate step-by-step design generation, to address the inherent limitations of LLMs. 

\begin{figure*}[t]
    \centering
    \vspace{-1em}
    \includegraphics[width=1\linewidth]{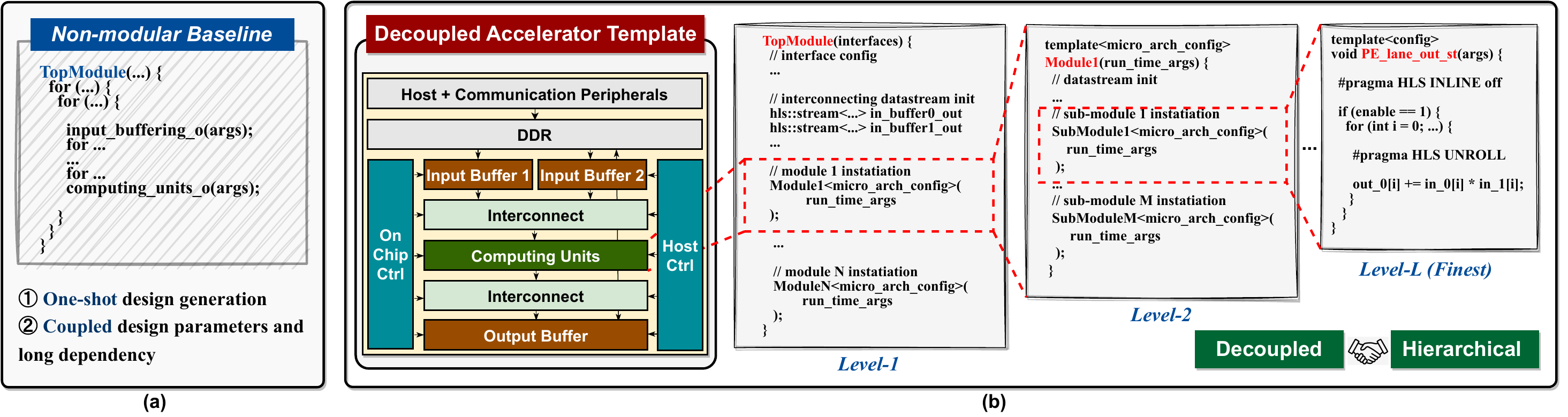}
    \vspace{-2.em}
   \caption{(a) LLMs with a non-modular template are limited by one-shot design generation, coupled design parameters, and long dependency; (b) In contrast, the proposed modular and decoupled accelerator template, which facilitates step-by-step design generation in a hierarchical manner.}
    \label{fig:template_overview}
    \vspace{-1em}
\end{figure*}

\begin{itemize}

    \item \textbf{High modularity}: Due to the token capacity restrictions of LLMs, the size of the input sample code used during in-context learning~\cite{razeghi2022impact,olsson2022context,chan2022data} and the code size for the final generated design during each round (i.e., a single LLM model inference) are significantly limited. With high modularity, the template is segmented into smaller and thus more manageable modules. Such a modular design generation approach can substantially reduce the required code size for both LLMs' input and output.

    \item \textbf{Decoupled module design}: Segmenting the code template into smaller modules can inadvertently introduce coupling and dependencies among their configuration settings. As discussed in Sec.~\ref{sec:failures}, this contradicts our goal of input token size reduction through high modularity, as the LLM must recall previous modules' settings. To counteract this, we propose independent module generation, each maintaining its own local settings. However, this can lead to sub-optimal overall designs, with potential data rate mismatches between connected modules causing stalls or deadlocks. To harmonize module operation, we suggest an additional search engine and adaptable inter-module communication schemes. These can optimally harmonize all local settings, mediating communication rates, and bandwidth discrepancies. Consequently, the LLM can generate each module based on its local settings, maintaining the decoupling principle.

    \item \textbf{Deep design hierarchies for step-by-step design generation}: The complexity of accelerators can result in large code sizes within even one single module, which may exceed the handling capacity of an LLM. To address this, our template adopts a hierarchy-based, module-by-module generation approach, streamlining the process and reducing the complexity at each stage. Every module consists of multiple sub-modules adhering to a decoupling principle, which may further contain their own sub-modules. This recursive nesting persists until further division is unfeasible (see Level-L in Fig.~\ref{fig:template_overview} (b)). This allows the LLM to systematically generate design hierarchies for each module, constraining the code size and complexity at each step.
\end{itemize}

\textbf{Overview of the proposed accelerator template.}
Incorporating the three key principles above, we introduce a new, modular, and decoupled accelerator micro-architecture and corresponding code template, as displayed in Fig.~\ref{fig:template_overview} (b). While here we focus on the widely-used GEMM operator considering its extensive application across various AI algorithms, our template retains a generic structure. Comprising a collection of versatile modules, each can be flexibly re-designed according to the prompts and local configuration settings, offering varying hardware efficiency or even unique functionalities. To guide precise and decoupled code generation by LLMs, each module in our template strictly corresponds to a function instantiation in the source code, as illustrated in Fig.~\ref{fig:template_overview} (b).
Each module is hierarchically composed of nested sub-modules to facilitate LLMs' step-by-step generation. Modules are interconnected via stream-based communication links and asynchronous data FIFOs, to reduce the controlling overhead of handling potential mismatches between different modules' data production and consumption rates. Processing onset and termination within each module primarily hinge on data availability, promoting fine-grained operation overlap and streamlining control overhead~\cite{khailany2018modular}. For modules with multiple input ports, such as the interconnect module in Fig.~\ref{fig:template_overview} (b), we include additional synchronization logic to ensure data alignment and accuracy.

\textbf{Key components of the proposed accelerator template.} We elaborate on different modules shown in Fig.~\ref{fig:template_overview} (b) below.
\begin{itemize}
    \item \textbf{Buffer modules}: These are designed to facilitate parallel data access for subsequent computing units and to harness varied data reuse patterns. They define (1) the on-chip memory partition and the corresponding data allocation for parallel access, and (2) the procedures for refreshing, resetting, or retaining the data within buffers, dictated by control signals associated with different data reuse patterns. Double buffers are assumed for all possible design styles to ensure optimal throughput.
    
    \item \textbf{Computing units modules}: These modules primarily handle computations, e.g., multiplications and additions, within their parallel computing units. Implemented as a collection of Multiply-and-Accumulate (MAC) units, their interconnects can be tailored according to different design prompts, striking a balance between spatial data reuse, MACs' data propagation latency, and on-chip buffer bandwidth contention. We structure nested design hierarchies for easier LLM-assisted generation and to accommodate a variety of MAC interconnect styles. For example, individual MACs may be linked to create a 1D PE-lane sub-module, while multiple PE lanes can be interconnected to forge a larger 2D PE-array module, enhancing scalability.

    \item \textbf{Interconnect modules}: They are designed to flexibly distribute and synchronize data between buffer modules and computing units modules. Their flexibility becomes crucial when the computing units consist of multiple 2D PE arrays and when the algorithm-to-PE-array mapping can change at run-time.

    \item \textbf{Control (Ctrl) modules}: They handle initial control data retrieval from the host, control data decoding, and potential runtime control data generation to alter various modules' modes.
    \item \textbf{Flexible communication arbitrators}: These are designed to manage potential mismatches in data production/consumption rates and bandwidths between interconnected modules, thus facilitating rate and bandwidth conversion.
\end{itemize}

\textbf{Implication and advantages of the proposed template.}
The proposed template offers three key benefits: \underline{(1)} by reducing the code size with decoupling module design and utilizing deep design hierarchies, the template allows LLMs to use limited input and output token capacity to generate complicated accelerator designs in a step-by-step manner, enhancing  LLMs' in-context learning capability;
\underline{(2)} as illustrated in Sec.~\ref{sec:finetuning}, LLMs' potential for generating AI accelerator designs can be further enlarged when finetuned using an additional dataset of sample code and prompt pairs. Our proposed template simplifies this finetuning process. It allows developers to generate the dataset in a module by module and hierarchy by hierarchy way, greatly reducing design complexity; and \underline{(3)} the principles identified for the proposed template extend beyond the domain of HLS. The same issues concerning LLM-assisted design generation, as outlined in Sec.~\ref{sec:failures}, still exist when leveraging LLMs to generate designs for other programming languages. Therefore, the aforementioned key principles are generally applicable, though the technical implementations for different domains may vary.

\vspace{-0.5em}
\subsection{The Demo-Augmented Prompt Generator Design}
\label{sec:prompt_design}
\vspace{-0.3em}

As outlined in Sec.~\ref{sec:insights} and prior studies \cite{bubeck2023sparks,gpt35,zhang2022opt}, a carefully enhanced prompt, reinforced with demonstrations, can effectively facilitate in-context learning for LLMs. This imparts LLMs with vital task-specific knowledge, unlocking their full capabilities. However, the prompt length limit makes it impractical to incorporate all possible demonstrations into one prompt. To tackle this, our demo-augmented prompt generator aims to efficiently generate prompts that automatically select the most relevant demonstrations from our crafted library and incorporate them into the prompt, balancing prompt length with the resulting in-context learning performance.

\textbf{Workflow.} Inspired by previous studies highlighting LLMs' ability in identifying similarities between different instructions~\cite{bubeck2023sparks}, we employ LLMs to facilitate demonstration selection within our demo-augmented prompt generator, which further minimizes the demand for human expertise in AI accelerator design. Specifically, as shown in Fig.~\ref{fig:framework}, in each iteration, given a design instruction generated by the search engine, we deploy an LLM to identify the similarity between the generated design instruction and those within the demonstration library. We then select the two most similar instructions, pairing them with their corresponding implementation as the demonstrations for this iteration of code generation. Then, we generate the demo-augmented prompt using the following template:

\begin{itemize}
    \item Assume you are an expert in HLS codes for AI accelerator design, I will now provide you with an instruction on generating the [\textit{Module Name}] for AI accelerator design. Below are two demonstration instructions and the corresponding generated code. Demo A: Instruction: [\textit{Demo A Instruction}]. Code: [\textit{Demo A Code}]; Demo B: Instruction: [\textit{Demo B Instruction}]. Code: [\textit{Demo B Code}]. Now please generate the code with the following instruction: [\textit{Design Instruction}]. 
\end{itemize}

\textbf{Demonstration library.} The high-quality demonstration library is a critical component in our demo-augmented prompt generator. The demonstrations within this library essentially serve as the primary knowledge source for LLMs to acquire domain-specific insights. As suggested in Sec.~\ref{sec:insights}, our goal is to assemble a demonstration library that encompasses a diverse range of design choices for the target domain, i.e., GEMM in this paper.
This ensures that the library provides demonstrations with abundant domain knowledge, catering to the diverse design instructions generated from the search engine. To achieve this, we adhere to the following guiding principles during the construction of our demonstration library:

\begin{itemize}
    \item \textbf{Highly correlated instruction and code pairs}: Each demonstration includes a detailed implementation instruction and the corresponding code, accompanied by comments. Each line of instruction is explicitly linked to specific code segments, providing clarity on their correlation and rationale.
    \item \textbf{Diverse design choices}: To ensure LLMs find demonstrations with sufficient domain knowledge for a given design instruction, we generate separate demonstrations illustrating modifications for each design parameter in our search space (Sec.~\ref{sec:other_funcs}).
\end{itemize}

Given the large search space size that GPT4AIGChip considers, generating a large number of distinct AI accelerator designs based on the principles outlined above requires substantial human effort. Fortunately, our proposed LLM-friendly hardware template (Sec.~\ref{sec:llm_friendly_hw_template}) makes it feasible. This template exhibits high modularity and allows for the decoupled  generation of each module within an AI accelerator. This approach significantly reduces the required human effort in two ways: \underline{(1)} each module is a concise and structured implementation focused on a specific function. Therefore, the implementation of one module does not require considering the impact on other modules, resulting in a streamlined and less labor-intensive implementation process; and \underline{(2)} not all modules in the template are affected by all the parameters in the search space. As a result, the potential design variations for each module are substantially reduced compared to the entire AI accelerator design.

\vspace{-0.3em}
\subsection{Implementation of Other Components in GPT4AIGChip}
\label{sec:other_funcs}
\vspace{-0.3em}

\textbf{Hardware design space.}
To ensure the performance of generated accelerators for diverse designs, a generic accelerator design space is crucial. It enables a flexible design process, providing the code generator with multiple options to customize the design of each target operator. Utilizing the template from Sec.~\ref{sec:llm_friendly_hw_template}, we identify five key hardware design parameters:

\begin{itemize}

    \item \textbf{MAC array sizes}: They indicate the total number of MACs in the MAC array of the instantiated accelerator.
    
    \item \textbf{Network-on-Chip (NoC) styles}: They determine the distribution of data to and from computing units, as well as the propagation of data among them. They can be classified into three main types: uni-cast, multi-cast, and broadcast. To promote design diversity, these styles are independently employed in various hierarchies of computing units, including individual MACs, 1D MAC lanes, and 2D MAC arrays. Furthermore, NoC styles are configured separately for different data types, e.g., the two input operands and one output in GEMM, enlarging design variations.

    \item \textbf{On-chip buffer sizes}: They denote the capacities of three principal buffers integral to the accelerator design, which consists of two input buffers and one (partial) output buffer. The sizes of all additional auxiliary buffers and registers are determined based on the capacities of these three principal buffers.
    
    \item \textbf{On-chip buffer partition styles}: They determine the data allocation among on-chip memory blocks within each buffer. By dividing the data across multiple blocks, parallel accesses can be achieved through buffer partitioning. Two dimensions, namely data width and height, can be employed for such partitioning.

    \item \textbf{Data reuse patterns}: They define how data once buffered is reused during computation. Changing how buffers and DRAM exchange data leads to diverse reuse strategies, e.g., first-input operand reuse, second-input operand reuse, or output reuse.
\end{itemize}

\textbf{The adopted search algorithm.}
GPT4AIGChip's accelerator search adopts an evolutionary algorithm known as tournament selection~\cite{miller1995genetic}, which iteratively evolves the accelerator's design. This iterative evolution process begins by initializing a population $P$, consisting of $|P|$ randomly selected accelerator designs $\{hw\}$ from the available design space.
Specifically, in each iteration, a subset $S$ of a fixed size is randomly selected from $P$. The superior designs, judged by top-tier hardware performance from LLM-generated implementation, become parent designs $\{hw\}_{parent}$. New accelerator designs $\{hw\}_{child}$ then emerge via mutation - a random tweak in a parent design parameter - and crossover - a random element exchange between two parents. The resultant $\{hw\}_{child}$ are added into $P$. To preserve a constant population size $|P|$, the oldest designs are phased out from $P$, following~\cite{real2019regularized}. Finally, once the maximum number of cycles is reached, the accelerator design with the highest performance throughout the search is chosen as the optimal solution.

\textbf{Design validation and code correction flow.}
GPT4AIGChip also integrates a process to validate and ensure the functionality of LLM-generated designs, consisting of three main stages: (1) Synthesizability Evaluation, (2) Correctness Verification, and (3) Performance Analysis.
In Synthesizability Evaluation, we initially employ standardized Vivado HLS tools~\cite{vivado_HLS} to synthesize LLM-generated codes. Subsequently, output log messages are processed via a custom error parser, armed with empirically-tested error detection and correction protocols. If the parser confronts errors beyond its capabilities, it may necessitate (1) LLM-driven design regeneration or (2) human intervention for error rectification. The current setup incorporates procedures for detecting and addressing undefined variables, improper HLS pragma usage, and out-of-bound array (memory) access.
Furthermore, we ensure that the design produces the desired results through the results correctness check. Specifically, a test-bench template is constructed featuring anticipated input and output. The produced output is subsequently contrasted with the expected output to confirm accuracy. Given the variety of possible errors, this step doesn't include automatic correction. In case of incorrect results, design regeneration is required.
Finally, we assess performance metrics (e.g., latency) and resource usage to provide feedback to the search engine, as depicted in Fig.~\ref{fig:framework}. Vivado HLS built-in tools generate these performance and resource estimations after design synthesis.

%% file: Sections/4-Experiment.tex
\section{Experimental Evaluation}
\label{sec:experiment}

\subsection{Experiment Setup}

We demonstrate the effectiveness of our GPT4AIGChip framework for automatically exploring the optimal AI accelerator design written in HLS on top of the design space introduced in Sec.~\ref{sec:other_funcs}. In particular, we adopt GPT-4~\cite{gpt4} as the default LLM and adopt the standard Vivado HLS design flow~\cite{vivado_HLS}, using a ZCU104 FPGA with XCZU7EV MPSoC~\cite{MPSoC}, for accelerator validation. We allocate 1024 digital signal processors (DSPs) for both baselines and our generated designs, unless noted. The final performance is measured onboard with PYNQ~\cite{pynq} as the deployment environment.

\vspace{-0.3em}
\subsection{Benchmark with Manual and Automated Designs}
\vspace{-0.3em}
To validate the quality of the generated hardware designs of our GPT4AIGChip framework, we benchmark with two baselines, including (1) hardware designs delivered by an industry-level design automation tool CHaiDNN~\cite{chaidnn_v2} and (2) manually optimized designs by human experts based on our template, in terms of the acceleration efficiency for six different networks under two input resolutions.  In the case of manually optimized designs, an expert hardware designer adjusts and finetunes the parameters outlined in our suggested templates. This process continues until either no further performance gains can be achieved within a reasonable timeframe (approximately one day) or the expert designer determines that no further improvements can be made based on empirical evidence.

\begin{figure}[t]
    \centering
    \includegraphics[width=0.98\linewidth]{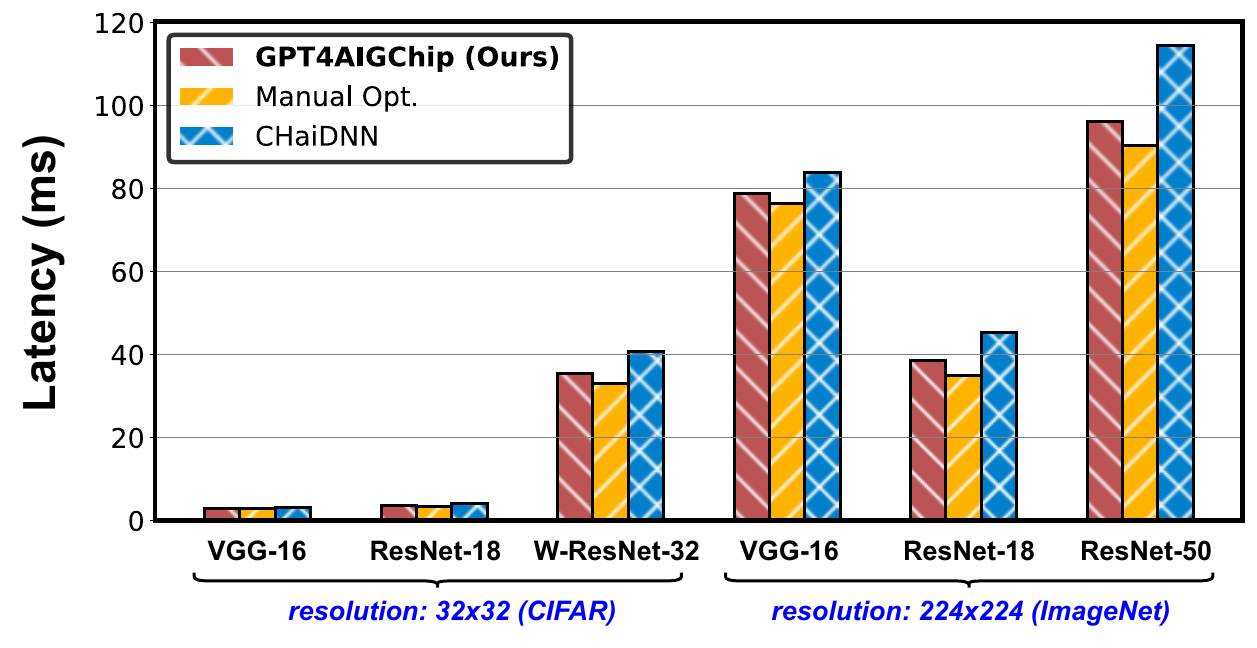}
    \vspace{-1em}
    \caption{Benchmark GPT4AIGChip's generated designs with CHaiDNN's~\cite{chaidnn_v2} generated ones and manually optimized ones by experts on six networks.}
    \label{fig:benchmark_sota}
    \vspace{-0.7em}
\end{figure}

\textbf{Observation.}
As shown in Fig.~\ref{fig:benchmark_sota}, we observe that \underline{(1)} our GPT4AIGChip framework can consistently outperform CHaiDNN~\cite{chaidnn_v2}, e.g., a 2.0\%$\sim$16.0\% latency reduction across six networks and two resolution settings, indicating that our GPT4AIChip has successfully and practically demonstrated the first LLM-powered AI accelerator design automation framework; and \underline{(2)} our searched accelerator design can match the hardware efficiency achieved by manual designs from human experts while enjoying much-reduced labor costs. This also implies that our searched accelerator implementation has the potential to serve as a good reference or initialization for human designers.

\begin{figure}[t]
    \centering
    \includegraphics[width=0.98\linewidth]{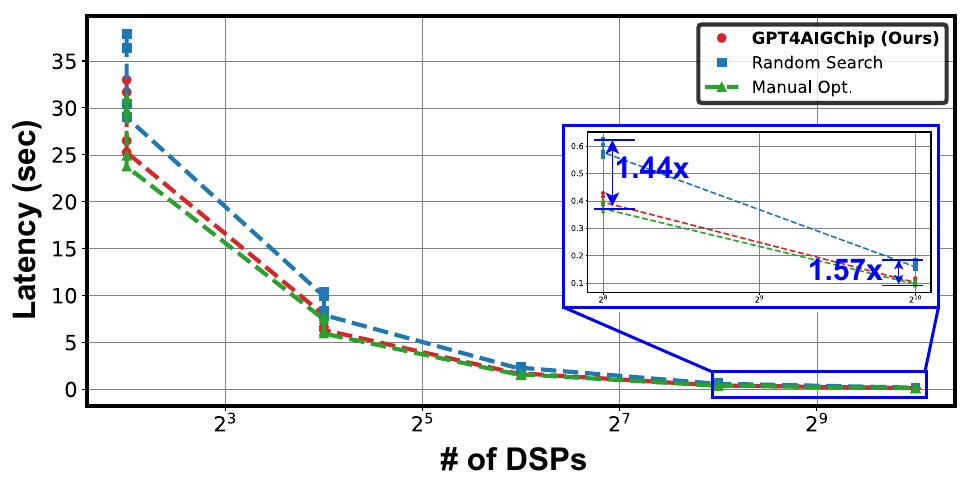}
    \vspace{-1em}
    \caption{Benchmark the latency-resource Pareto frontier achieved by GPT4AIGChip's search scheme, random search, and manual optimization.}
    \label{fig:search_scheme}
    \vspace{-1em}
\end{figure}

\vspace{-0.3em}
\subsection{Effectiveness of the Search Scheme}
\vspace{-0.3em}
To validate the effectiveness of our search scheme, we visualize the Pareto frontier, i.e., the trade-off between latency and the utilized resources in terms of DSPs, on top of ResNet-50 with an input resolution of $224\times224$, achieved by the hardware designs searched via GPT4AIGChip and benchmark with those delivered via a random search in our design space. In addition, we also benchmark with manually optimized accelerator implementation by human experts. 

\textbf{Observation.}
In Fig.~\ref{fig:search_scheme}, we can observe that although the designs generated by a random search under-perform manually optimized designs by human experts in terms of the achievable latency-resource trade-off under a relatively rich DSP setting, our GPT4AIGChip framework can outperform the random search baseline and match the hardware efficiency achieved by manual expert designs under comparable resources,
indicating the effectiveness of our search scheme in Sec.~\ref{sec:other_funcs}. Note that given a tight DSP budget, the efficiency of different designs is naturally similar.

\begin{table}[t]
    \centering
    \caption{Ablation study on the prompt format and explicitness.}
    \vspace{-0.5em}
    \resizebox{0.48\textwidth}{!}{
    \begin{tabular}{c|ccc}
    \toprule
        Prompt & \textit{No Demo} & \textit{High-Level Description} & \textit{\textbf{Demo-Augmented (Ours)}}  \\
        \midrule
        Pass@10 & 10\% & 30\% & 60\% \\
        \bottomrule
    \end{tabular}
    }
    \vspace{-0.7em}
    \label{tab:prompt design}
\end{table}

\begin{table}[t]
    \centering
    \caption{Ablation study on the demonstration selection criteria.} 
    \vspace{-0.5em}
    \begin{tabular}{c|ccc}
    \toprule
        Criteria & \textit{Dissimilar} & \textit{Random} & \textit{\textbf{Similar (Ours)}}  \\
        \midrule
        Pass@10 & 30\% & 50\% & 60\% \\
        \bottomrule
    \end{tabular}
    \vspace{-1.2em}
    \label{tab:demo select}
\end{table}

\vspace{-0.3em}
\subsection{Ablation Study of the Prompt Generator}
\vspace{-0.3em}
To validate the effectiveness of our proposed demo-augmented prompt generator, we validate the impact of different prompt design strategies on LLMs' accelerator design generation performance. Specifically, we use the generation of the output stationary computing units defined in Sec.~\ref{sec:llm_friendly_hw_template} as the target application with Pass@10, as defined in Sec.~\ref{sec:finetuning}, as the evaluation metric.

\textbf{Prompt format and explicitness.} We first validate the necessity of using the demo-augmented prompt when generating the code. Specifically, we consider two alternative prompt designs: (1) \textit{No Demo}, which uses the same design instruction but without demo instruction and code; (2) \textit{High-Level Description}, which replaces the explicit instructions in the demo with high-level descriptions (e.g., output stationary computing units). As shown in Tab.~\ref{tab:prompt design}, our demo-augmented prompt improves the LLM's ability in generating desired accelerator designs by 50\% and 30\% in terms of Pass@10 rates over the two baselines \textit{No Demo} and \textit{High-Level Description}, respectively.

\textbf{Demonstration selection.}
We further validate the importance of incorporating appropriate demonstrations in the prompt. Specifically, we consider two alternatives other than selecting the most similar demonstrations: (1) random selection, denoted as \textit{Random}; (2) selecting the most unrelated demonstrations, denoted as \textit{Dissimilar}. As shown in Tab.~\ref{tab:demo select}, using the most similar demonstrations can better impart necessary knowledge in the prompt, leading to 10\% and 30\% higher Pass@10 rates over \textit{Random} and \textit{Dissimilar}, respectively.

\textbf{The number of demonstrations.}
We also evaluate the number of demonstrations needed in the prompt. Tab.~\ref{tab:prompt number} shows that \underline{(1)} introducing demonstration (i.e., \# Demo $>$ 0) significantly improves the performance, and \underline{(2)} providing 2 demonstrations leads to a satisfactory trade-off between Pass@10 rates and prompt length. 

\begin{table}[h]
    \vspace{-1em}
    \centering
    \caption{Ablation study on the number of demonstrations provided. The setting adopted by GPT4AIGChip is \textbf{bolded}.}
    \vspace{-0.5em}
    \begin{tabular}{c|cccc}
    \toprule
        \# Demo & \textit{0} & \textit{1} & \textit{\textbf{2}} & \textit{3} \\
        \midrule
        Pass@10 & 10\% & 50\% & 60\% & 60\% \\
        \bottomrule
    \end{tabular}
    \vspace{-1em}
    \label{tab:prompt number}
\end{table}

\begin{figure*}[t]
    \centering
    \vspace{-0.5em}
    \includegraphics[width=0.98\linewidth]{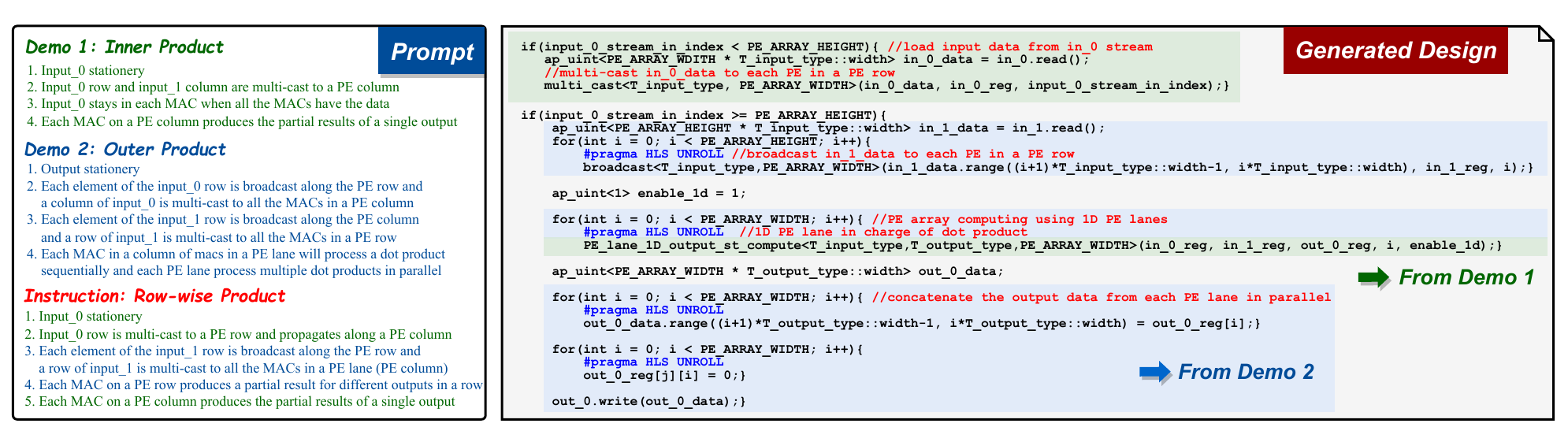}
    \vspace{-0.7em}
    \caption{Visualize the generated row-wise-product-based implementation of GPT4AIGChip following the instructions and the demonstrations of inner-/outer-product-based matrix multiplication, for which we show the design descriptions only and hide the codes for visual clarity.}
    \label{fig:code_visual}
    \vspace{-1.3em}
\end{figure*}

\vspace{-0.3em}
\subsection{Visualizing the Generated Designs}
\vspace{-0.3em}
\label{sec:code_visual}
To better illustrate the design generation process from in-context demonstrations, 
we provide one example in Fig.~\ref{fig:code_visual}, where the LLM is prompted by two demonstrations that correspond to the inner-/outer-product-based matrix multiplication implementation, respectively, and is instructed to generate the implementation of a row-wise-product-based design. We can observe that our framework exhibits good logical reasoning about the relationship between the desired design and the two given demonstrations, as the generated implementation can successfully identify the reusable codes from the two demonstrations and fuse them in a logically correct manner, i.e., adopt a spatial reduction along the PE lane, following the inner-product solution, and broadcast one element along the PE row, following the outer-product solution.
This indicates the generalization capability of our GPT4AIGChip framework.

%% file: Sections/5-Envision.tex
\section{Limitations and Future Work}
\label{sec:envision}

We recognize that our work marks the initial step towards LLM-aided AI accelerator design automation, with significant efforts still needed to propel this promising yet demanding field forward. In order to provide insights into future research within this domain, we discuss our limitations and the resulting inspiration for future work.

\textbf{Dependence on the demonstration library.} Our framework's dependence on an available demonstration library tailored to the target domain is still essential to compensate for the existing LLMs' limited comprehension of AI accelerator designs. However, constructing such a library demands non-trivial hardware expertise and constrains the capacity for cross-domain applicability. To tackle this challenge, a promising avenue for future investigation involves amassing annotated hardware design codes alongside corresponding descriptions, which can be utilized for LLM finetuning. This approach encodes AI accelerator design knowledge into the LLMs' weights, diminishing the reliance on human expertise and enhancing generalization across diverse domains. Additionally, the accumulation of high-quality paired data can also set a benchmark for the community to evaluate the capabilities of forthcoming LLMs.

\textbf{Human involvement requirements.} Our framework still necessitates human involvement to rectify the interfaces of the generated hardware modules and assemble them into the final accelerator. This requirement arises from the diverse implementation styles of different design choices, leading to inevitable inconsistencies in the module-wise interface, particularly when expanding the hardware template to encompass a wider array of domains. In future endeavors, this challenge could potentially be addressed by crafting novel hardware templates and design principles that strike a balance between generality and modularity. Such an approach would cater to the preferred formats of both human designers and LLMs, potentially reducing the need for extensive human intervention. Additionally, when coupled with LLM finetuning, explicit regularizations can be applied to LLMs to ensure the consistency of interface implementation.

\textbf{Verification cost of the generated accelerator.} Beyond the accelerator design phase, the verification cost associated with accelerator design can be substantial, a facet not encompassed by our present framework. Addressing this omission and consequently furnishing a comprehensive and accurate AI accelerator generation pipeline entails the development of verification tools tailored to LLM-generated accelerator designs. In future endeavors, such tools are of paramount importance to guarantee the soundness of the designs and to automatically rectify prevalent LLM-related anomalies. These tools could be constructed based on human insights into LLM design biases and/or through the creation of another LLM specifically customized for verification purposes.

%% file: Sections/6-Related-Work.tex
\section{Related Work }

\textbf{LLMs and in-context learning.}
LLMs, e.g., GPT-4~\cite{gpt4} and Alpaca~\cite{koubaa2023gpt}, have demonstrated outstanding performance across various tasks, attributed to extensive pretraining, high-quality datasets, and effective tuning methods~\cite{raffel2020exploring,devlin2018bert,koubaa2023gpt,zhang2019dialogpt,floridi2020gpt,radford2018improving,radford2019language,gpt4}. One key ability of LLMs is in-context learning, where providing a few relevant input-output pairs can largely enhance task performance~\cite{zhang2022active,garg2022can,zhao2021calibrate,min2021noisy,liu2021makes,rong2021extrapolating,brown2020language,razeghi2022impact}.
Nevertheless, employing in-context learning for AI accelerator design automation encounters challenges stemming from token length limitations in LLMs (e.g., 4096 tokens for GPT-4~\cite{gpt4}). The complexity of AI accelerator design code can easily surpass this limit, necessitating a decoupled hardware template and the judicious selection of highly correlated demonstrations.

\textbf{LLM-powered code generation.}
Previous work has leveraged LLMs to generate code, particularly in languages with substantial pretraining resources like Python, or highly structured ones like SQL~\cite{poesia2022synchromesh,chen2023improving,nijkamp2022codegen,wang2022code4struct,zheng2023codegeex}. For hardware code generation, LLMs have been finetuned on Verilog codes for code completion~\cite{thakur2022benchmarking} and used for bug-fixing~\cite{ahmad2023fixing}. 
Existing LLM-powered code generation primarily targets languages that are highly structured and akin to natural languages or focuses on simpler tasks such as code completion. They are not able to generate complex, domain-specific hardware designs based on human instructions, which require a deep understanding and reasoning of the target domain. 
In contrast, our work represents the first endeavor to harness the power of LLMs for generating AI accelerator designs based on natural languages.

\textbf{Design automation for DNN accelerators.}
To facilitate AI accelerator design, various design automation tools have been developed for FPGA and/or ASIC-based accelerators. For example, Deepburning~\cite{wang2016deepburning} and TensorLib~\cite{jia2021tensorlib} utilize pre-built templates with customizable parameters to construct FPGA-based DNN accelerators; 
AutoDNNChip~\cite{Xu_2020} adopts a cohesive graph representation, encompassing a broad spectrum of accelerator designs, to enhance the applicability of its proposed design automation. However, these existing tools still require nontrivial hardware expertise and often rely on domain-specific languages, necessitating deep knowledge of hardware architecture and thus limiting the accessibility to most AI developers. To address these limitations, GPT4AIGChip aspires to produce high-quality hardware codes in response to natural language instructions, reducing the entry barrier for AI developers unfamiliar with hardware and thus expediting AI accelerator innovations.

%% file: Sections/7-Conclusion.tex
\section{Conclusion }

This work delves into the capabilities of LLMs for AI accelerator design automation. As a crucial first step, we conducted an in-depth investigation of LLMs' strengths and limitations in automated AI accelerator generation, drawing vital insights into the prospects of LLM-powered design automation. Building on these insights, we develop GPT4AIGChip, which integrates an automated prompt-generation pipeline using in-context learning to guide LLMs towards creating high-quality AI accelerator designs. Various experiments and ablation studies validate the effectiveness of GPT4AIGChip in generating high-performance AI accelerators in response to human natural languages. To our knowledge, this work marks the first successful demonstration of a pipeline for LLM-powered automated AI accelerator generation, highlighting the untapped potential of LLMs in design automation and suggesting promising avenues for next-generation AI accelerator development.

\section*{Acknowledgement}
The work is supported by the National Science Foundation (NSF)
through the CCRI program (Award number: 2016727), an NSF
CAREER award (Award number: 2048183), and CoCoSys, one of the seven centers in JUMP 2.0, a Semiconductor Research Corporation (SRC) program sponsored by DARPA.